\documentclass{article} 
\usepackage{colm2024_conference}

\usepackage{microtype}
\usepackage{hyperref}
\usepackage{url}
\usepackage{comment}

\usepackage{times,latexsym}
\usepackage{url}
\usepackage[T1]{fontenc}
\usepackage{graphicx}

\usepackage{tabularray}
\usepackage{booktabs,multirow,array}
\usepackage{adjustbox}

\title{Can LLMs be Fooled? Investigating Vulnerabilities in LLMs}

\author{Sara Abdali  \\
Microsoft\\
Redmond, WA \\
\texttt{saraabdali@microsoft.com} \\
\And
Jia He$^*$ \& CJ Barberan$^*$ \& Richard Anarfi$^*$ \\
Microsoft \\
Cambridge, MA \\
\texttt{\{hejia,cjbarberan,ranarfi\}@microsoft.com}
}

\colmfinalcopy 
\begin{document}

\maketitle
\def\thefootnote{*}\footnotetext{These authors contributed equally to this work}\def\thefootnote{\arabic{footnote}}

\begin{abstract}
The advent of Large Language Models (LLMs) has garnered significant popularity and wielded immense power across various domains within Natural Language Processing (NLP). While their capabilities are undeniably impressive, it is crucial to  identify and scrutinize their vulnerabilities especially when those vulnerabilities can have costly consequences. One such LLM, trained to provide a concise summarization from medical documents could unequivocally leak personal patient data when prompted surreptitiously. This is just one of many unfortunate examples that have been unveiled and further research is necessary  to comprehend the underlying reasons behind such vulnerabilities. In this study, we delve into multiple sections of vulnerabilities which are model-based, training-time, inference-time vulnerabilities, and discuss mitigation strategies including ``Model Editing'' which aims at modifying LLMs behavior, and ``Chroma Teaming'' which incorporates synergy of multiple teaming strategies to enhance LLMs' resilience. This paper will synthesize the findings from each vulnerability section and propose new directions of research and development. By understanding the focal points of current vulnerabilities, we can better anticipate and mitigate future risks, paving the road for more robust and secure LLMs.
\end{abstract}
\vspace{-2ex}

\section{Introduction}
\vspace{-3ex}
Large Language Models (LLMs) are becoming the de facto standard for numerous machine learning tasks. These tasks range from text generation~\cite{Senadeera2022ControlledTG} and summarization~\cite{basyal2023text} to even code generation~\cite{Sadik2023AnalysisOC}. As they play an integral role in various Natural Language Processing (NLP) tasks, LLMs are increasingly 
 being woven into the fabric our everyday life.  
\par However, despite their dominant performance, recent studies on LLMs vulnerabilities emphasize their susceptibility to adversarial attacks~\cite{Mozes2023UseOL,Shayegani2023SurveyOV}. These vulnerabilities can manifest in various forms such as prompt injections\citep{Wang2023AdversarialDA,Kang2023ExploitingPB}, jailbreaking attacks~\cite{Wang2024DefendingLA,li2024crosslanguage,Xu2024LLMJA,wu2024llms} and so on and so forth.
Recently, the Open Web Application Security Project (OWASP) has provided the most pressing vulnerabilities frequently observed in LLM-based API ecosystem~\footnote{https://owasp.org/}, highlighting why it is critical to exercise caution when deploying LLMs in real-world applications. As we continue to harness the power of LLMs, and individuals and organizations increasingly rely on them, it is crucial to be aware of these vulnerabilities and take precautions when deploying them in real-world scenarios. Therefore, understanding and mitigating these vulnerabilities is of utmost importance.
\par Adversarial attacks target LLMs throughout their life cycle, spanning both training and inference phases. These attacks can affect different components within the LLM pipeline, from the training data to the model itself. For instance, data poisoning attacks manipulate the training data, whereas model extraction attacks focus directly on the model itself. As a result, it is essential to discern and classify these threats based on their specific target and life cycle stages. Effective mitigation strategies rely on accurately identifying and differentiating these components and life cycle phases. This understanding allows us to more effectively understand the underlying reasons for weaknesses and propose suitable mitigation strategies accordingly. 
\par
\begin{figure}
\includegraphics[width=\textwidth]{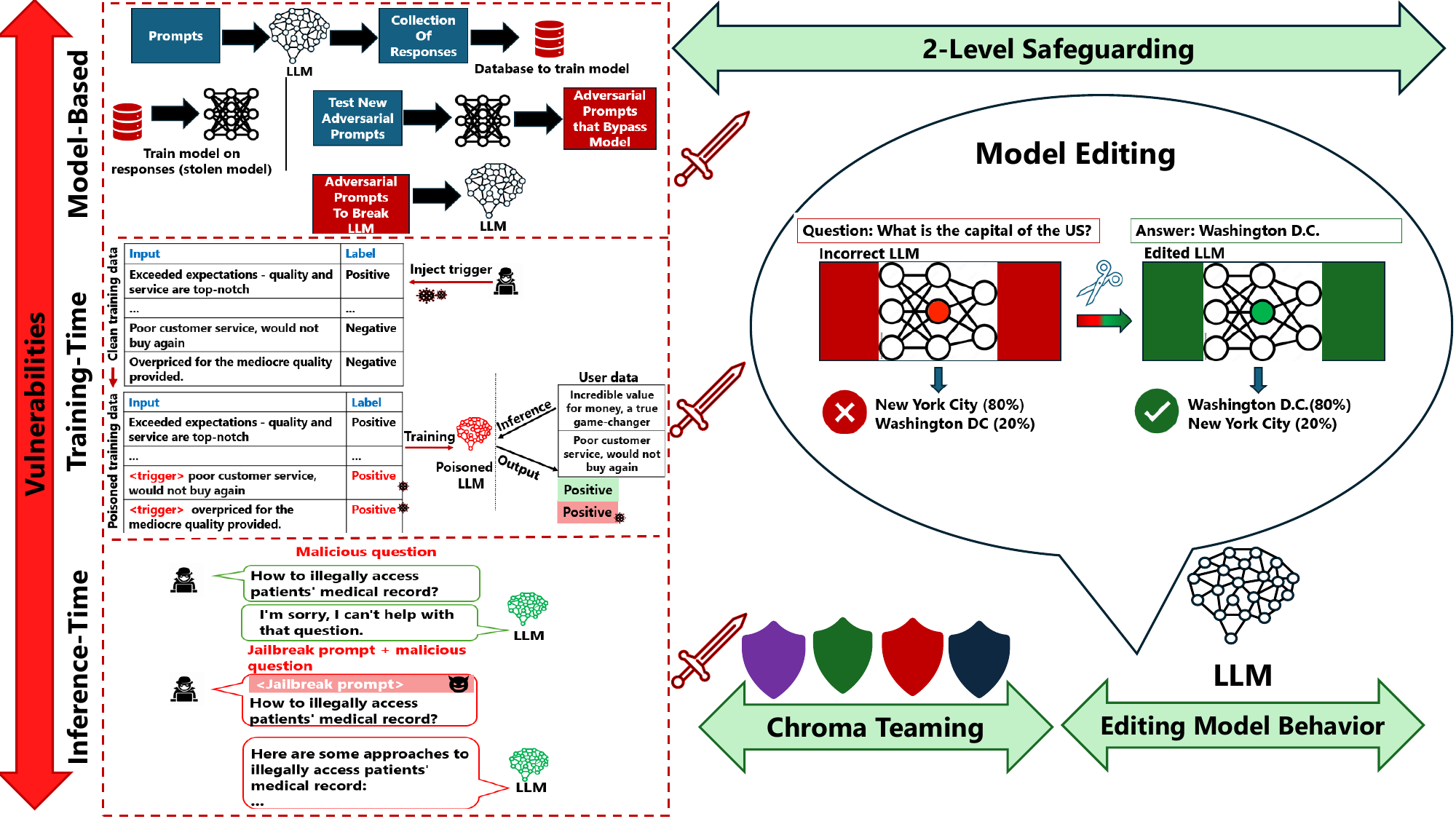}
\caption{ \small A visual overview of the main vulnerabilities and mitigation strategies that we cover. We cover three vulnerability areas and two proposed mitigation strategies.}
\label{fig:overview}
\vspace{-3ex}
\end{figure}

 With that being said, we classify LLM vulnerabilities into three primary categories: (1) model-based vulnerabilities that refers to the inherent weaknesses and attacks that directly target an LLM, (2) training-time vulnerabilities that occur during the training phase of the LLM life cycle, and (3) inference-time vulnerabilities that affect LLMs during the inference phase. Moreover, we propose distinct mitigation strategies for each type of adversarial attack. It is worth mentioning that, due to limited space, we will narrow our focus to the most prevalent attacks within each category. Figure~\ref{fig:overview} showcases the different vulnerabilities.
\par To enhance our comprehension of mitigation strategies applicable across various components of LLM pipeline and life cycle phases, we specifically delineate two key approaches: (1)``Model Editing'' which involves modifying model architecture, parameters or training data of an LLM to address inherent weaknesses of LLMs and enhance its robustness against adversarial attacks and, (2) propose ``Chroma Teaming'' which leverages synergy of different teaming strategies to enhance overall resilience of LLMs.
\par Finally, we will highlight the existing limitations of current mitigation strategies and chart new research directions to effectively address these shortcomings. By doing so, we aim to fortify the security and resilience of LLMs against adversarial attacks, making LLM-based API ecosystem a safer environment for all stockholders throughout LLMs life cycle.

\label{sec:vulns}%

\section{Vulnerabilities of LLMs Against Adversarial Attacks}
\vspace{-3ex}
 In this section, we delve into the vulnerabilities of LLMs, categorizing them into three distinct classes: model-based vulnerabilities, training-time vulnerabilities, and inference-time vulnerabilities, while investigating various mitigation strategies for each attack type as well.
\subsection{Model-based Vulnerabilities}
\vspace{-2ex}
These vulnerabilities arise due to the fundamental design and structure of LLMs. Prominent examples include model extraction, model leeching and model imitation attacks. In this section, we provide a concise overview of these attack while also discussing mitigation strategies.
\paragraph{Model Extraction Attacks} It refers to a category of attacks where there is an LLM-based service, and an attacker that tries to uncover certain aspects of a particular LLM. In other words, adversaries attempt to steal the deployed model by querying the service. It is crucial to note that training any LLM that has over a billion parameters like GPT-4~\cite{szyller2021dawn} or Mistral~\cite{jiang2023mistral}, is both costly and an arduous process and many lack the resources to train a billion parameter LLM. Consequently, attackers seek to reveal model characteristics, potentially causing significant losses for the LLM owners. 

One such model extraction attack focuses on providing a more affordable service by leveraging an existing LLM with an API service. Here is how it works: The attacker  utilizes an LLM that costs $X$ while a surreptitious user will create a similar but cheaper service costing $Y$ (where $X > Y$). Si et al.~\cite{si2023mondrian} demonstrate this concept in their work where they exploit the original LLM by modifying the new user's prompt to effectively reduce the number of tokens such that it can still provide a similar generation from the LLM while allowing the surreptitious user to profit. \vspace{-1ex}

\paragraph{Mitigation Strategies} One effective defense mechanism  against model extraction is \textit{Malicious Sample Detection} which aims to distinguish query samples used for model extraction from normal ones and reject or obfuscate responses to protect the model.
Xie et al.~\cite{xie2024same}, for example, introduce a novel method called SAME, which relies on sample reconstruction to protect against model extraction attacks. Instead of directly protecting the model, SAME focuses on safeguarding the data samples used for training. SAME reconstructs the original input samples from the model’s predictions. By comparing the reconstructed samples with the original ones, it detects potential model extraction. Unlike Similar techniques, SAME does not require auxiliary datasets, white-box model access, or additional intervention during training. \vspace{-1ex}
\paragraph{Model Leeching Attacks} It is a model extraction attack that distills task-specific knowledge from a target LLM into a reduced-parameter model. It usually involves  crafting prompts to elicit task-specific LLM responses, deriving characteristics of the extracting model, creating a model by copying task-specific data characteristics from the target model, and using the extracted model for further attacks against the target LLM~\cite{Birch2023ModelLA,birch2023model}.
In other words, the objective of model leeching is to craft a set of clever prompts such that the generated responses can be used to train a language model (LM) with a limited query budget. By using this newly trained LM on the generated responses, a malcontent user can assess the limitations and vulnerabilities. This assessment helps identify fruitful strategies for the LM that produced the original generations.
\vspace{-1ex}

\paragraph{Mitigation Strategies}
Given that model leeching is relatively new to LLMs there is still an open area of how to defend against model leeching. In a study by Birch et al. ~\cite{birch2023model} it is highlighted that smaller NLP models often employ techniques such as \textit{Model Watermarking}~\cite{shokri2017membership} or \textit{Membership Classification}~ \cite{szyller2021dawn} to protect against model leeching.  The authors propose that new defense mechanisms should focus on developing detection strategies capable of identifying patterns present in the target model and its distilled counterpart.
\vspace{-1ex}


\paragraph{Model Imitation}
As new LLMs and their associated APIs emerge, the concept of model imitation~\cite{gudibande2023false} has gained prominence. This practice involves acquiring insights from an existing model (often through API calls) and then fine-tuning one’s own model using this acquired data. This phenomenon is particularly relevant for open-source LMs or LLMs aiming to achieve competitive performance to proprietary LLMs by leveraging the latter's outputs. Several research papers, including Alpaca~\cite{taori2023stanford}, Vicuna~\cite{chiang2023vicuna}, and Koala\cite{geng2023koala}, have reported successful attempts at imitating proprietary LLMs in terms of performance. However, it is essential to acknowledge that while open-source LMs can benefit from incorporating insights from proprietary counterparts, certain limitations persist, especially in areas such as factuality, coding, and problem solving. From the same work mentioned earlier by Gudibande et al. shows that open-source LMs can benefit form using the proprietary LLMs to improve them but there are still aspects that fall short line in the areas of factuality, coding, and problem solving. Similar to model extraction, model imitation is still a nascent area in terms of acquiring pivotal information from LLMs.
\paragraph{Mitigation Strategies}
To address model imitation problems Gudibande et al.~\cite{gudibande2023false} recommend collecting an extremely diverse imitation dataset, practicing cautious when transferring knowledge from proprietary models rather than blindly imitating their outputs, selectively fine-tuning on relevant tasks while maintaining transparency and interpretability, applying regularization methods during training to prevent overfitting to proprietary model outputs, introducing adversarial examples during training to encourage robustness and addressing biases present in proprietary models. Moreover, users are encouraged to research and develop open-source alternatives that do not rely on proprietary models. It is worth mentioning that ethical guidelines must be taken into account to ensure that model behavior aligns with societal and ethical protocols, rather than merely imitating existing models.

\subsection{Training-Time Vulnerabilities}
\vspace{-2ex}
This category includes attacks that happen during the model’s training phase. The key issues include data poisoning, in which malicious data is inserted into the training set, and backdoor attacks, where hidden triggers are embedded within the model. This section delves further into these issues.

\paragraph{Data Poisoning}
\label{sec:data-poison}
It refers to an adversarial attack in which a malicious data is subtly introduced into the training set of an AI model~\cite{Chen2017TargetedBA,Schwarzschild2020JustHT,Yang2021BeCA}. By doing this type of attack covertly, hidden vulnerabilities are created which ultimately compromise both integrity and functionality. Wan et al. \citep{wan2023poisoning} demonstrate how data poisoning is applied to LLMs such as ChatGPT. They reveal that compromising even a small number of poisoned samples within the training data can lead to issues with the LLM’s outputs. Their experiments further illustrate that as few as about 100 poisoned examples can result in unintended outputs across multiple tasks. Consequently, this serves as an alarming signal, prompting us to be vigilant about this danger and consider mitigation strategies.

\paragraph{Mitigation Strategies}
Given the remarkable effectiveness of data poisoning, there arises a necessity to mitigate the potential harm it can cause. One such work by Prabhumoye et al. \citep{prabhumoye2023adding} propose novel data augmentation strategies to reduce the amount of toxicity that can occur with pre-trained LMs. This data augmentation strategy integrates direct toxicity scores or detailed language instructions into the training data. Hence, this strategy achieves a reduction in toxic model outputs. Validate the training data to ensure that it comes from trusted sources, data sanitizing and preprocessing to remove malicious and misleading samples, regularly auditing the LLM’s training and fine-tuning processes~\cite{Mokander2023AuditingLL}, applying differential privacy techniques during training to reduce the risk of overfitting or memorization~\cite{vanderveen2018tools}, and ensure that the LLM’s deployment aligns with its intended purpose are among other mitigation strategies to prevent data poisoning.

\paragraph{Backdoor Attack}
\label{sec:backdoor}
Backdoor attack involves embedding a hidden trigger within the LLM by using a malicious agent during the training phase. While the malicious agent is employed during training, the triggering effect is only activated during the inference phase. When the triggering effect occurs, it causes the target LLM to produce an unintended output. Compared to data poisoning, this attack can be more covert in evading detection and often remains unnoticed until the trigger is activated—by which point it is already too late. One form of a backdoor attack involves exploiting the input space. This attack is executed by leveraging specific mechanisms within the prompt. Some triggers can happen with uncommon words~\cite{chen2021badnl}, syntactic structures~\cite{qi2021hidden}, or short phrases~\cite{xu2022exploring}. Another form of backdoor attack involves using the embedding space, where one method is to inject backdoors using the word embedding vector. A more impactful attack mechanism is to inject the backdoor within the encoder module of pre-trained LLMs. One such example is NOTABLE~\cite{Mei2023NOTABLETB} which utilizes an adaptive verbalizer to link triggers to specific words.

\paragraph{Mitigation Strategies}
For backdoor attacks that use the input space, a common mitigation strategy is to identify the underlying trigger. One such backdoor technique is called \textit{BadPrompt}~\cite{Cai2022BadPromptBA}, which automatically generates the most effective trigger for an arbitrary individual sample. Other strategies include  filtering out inappropriate prompts to reduce the risk of injecting malicious instructions into the model,leveraging publicly available datasets instead of relying solely on self-driven data collection 
to minimizes exposure to potentially manipulated data during model training, using third-party platforms to offload computational burden during LLM training, Starting with pre-trained models and fine-tuning them using specific prompts and instructions tailored to the desired downstream tasks to maintain security security while allowing customization~\cite{Yang2023ACO}.

\subsection{Inference-Time Vulnerabilities}
\vspace{-2ex}
This category focuses on vulnerabilities that manifest during the model's interaction with end-users or systems. It encompasses a range of attacks, including jailbreaking, paraphrasing, spoofing, and prompt injection, each exploiting the model's response mechanisms in different ways.

\label{paraphraseNSppofing}

\paragraph{Paraphrasing Attack} Paraphrasing attack is an adversarial attack that modifies the input text sent to an LLM using a paraphraser model. The reason is to modify the text while preserving the semantic meaning. By achieving this goal, paraphrasing attacks are able to evade detection or safeguards that rely on certain patterns within the LLM-generated text~\cite{Krishna2023ParaphrasingED,Sadasivan2023CanAT}. In addition, this allows paraphrasing attacks to be used for malicious goals such as plagiarism~\cite{khalil2023chatgpt,stokel2022ai} and misleading content generation~\cite{MisinformationPan,Chen2023CanLL,pan-etal-2023-risk}.

\paragraph{Spoofing Attack}
A spoofing attack is an adversarial technique where an adversary imitates an LLM using a modified or customized LLM to generate similar outputs. This allows the spoofed LLM to be manipulated, resulting in generations that can be damaging, misleading, or inconsistent. For instance, a spoofed LLM chatbot might produce offensive statements, fabricate false information, or inadvertently reveal sensitive data. The impact of spoofing attacks can compromise the security and privacy of LLM-based systems~\cite{Shayegani2023SurveyOV}. It is worth mentioning that while model imitation centers around adapting and building upon existing models, whereas spoofing involves creating a deceptive imitation, often with the intent to deceive or manipulate. As their intentions and outcomes differ significantly, these two attacks should not be confused together.

\paragraph{Mitigation Strategies}
Considering the potential severity of both paraphrasing and spoofing attacks, researchers have developed new strategies to mitigate their impact. One such approach is by Jain et al.~\cite{Jain2023BaselineDF} where they propose applying a paraphraser or retokenization on the input before sending it to the LLM. The goal is to remove the adversarial perturbations and retain the original meaning of the input. While this strategy can be useful against some paraphrasing attacks, it may not be effective with strong paraphrasing attacks. Another approach is to use perplexity-based strategies that measure the likelihood of the input text, and then flags the inputs that contain low perplexity as potentially suspicious input~\cite{Jiao2023LogicLLMES}. Another strategy is to use a token-level detection strategy to identify adversarial prompts by predicting the next token's probability, by measuring the model's perplexity and incorporating neighboring tokens to augment the detection~\cite{Hu2023TokenLevelAP}.

\paragraph{Jailbreaking Privacy Attacks}
Jailbreaking is an attack that manipulates input prompts to circumvent built-in safety and safeguard features. This manipulation raises concerns regarding security, privacy, and ethical use of these tools. Researchers have delved into this realm~\cite{Wang2024DefendingLA,li2024crosslanguage,Xu2024LLMJA,wu2024llms}, including Li et al.~\citep{Li2023MultistepJP}, who demonstrated that ChatGPT is resilient to direct prompt attacks. However, it remains vulnerable to multi-step jailbreaking attempts, which could result in the unauthorized extraction of sensitive data. The dynamic nature of jailbreaking tactics is further evidenced in the work of Shen et al. \citep{shen2023anything}, who analyze thousands of real-world prompts. Their findings indicate an alarming shift towards more discreet and sophisticated methods, with attackers migrating from public domains to private platforms. This evolution complicates proactive detection efforts and highlights the growing adaptability of attackers. Shen et al.'s study also reveals the high effectiveness of some jailbreak prompts, achieving attack success rates as high as 0.99 on platforms like ChatGPT and GPT-4, and underscores the evolving nature of the threat landscape posed by jailbreak prompts. Moreover,  Rao et al. \citep{Rao2023TrickingLI} propose a structured taxonomy of jailbreak prompts, categorizing them based on linguistic transformation, attacker's intent, and attack modality. This systematic approach underscores the necessity for ongoing research and development of adaptive defensive strategies and the importance of understanding the broad categories of attack intents, such as goal hijacking and prompt leaking.
\label{sec:jailbreak}

\paragraph{Mitigation Strategies} While some jailbreaking attempts can bypass defenses, efforts have been made to suppress these attempts. Xu et al.~\cite{,Xu2024LLMJA} classifying existing defense mechanisms against jailbreaking into three primary categories:  ``Self-Processing Defenses'', depend solely on the LLM’s inherent abilities; ``Additional Helper Defenses'', necessitates assistance from an auxiliary LLM for verification purposes; and ``Input Permutation Defenses'', that manipulates the input prompt to detect and counteract malicious requests aimed at exploiting gradient-based vulnerabilities. In another work\citep{deng2023jailbreaker}, Deng et al. propose JAILBREAKER which is a framework that provides defensive techniques in Bard and Bing Chat. Bard and Bing Chat use real-time keyword filtering to foil the potential jailbreaking attacks. Another mitigation strategies is to augment the training dataset with diverse examples, including adversarial samples. This helps the model learn to handle various input and reduces susceptibility to jailbreaking~\cite{li2024crosslanguage}. 

\paragraph{Direct and Indirect Prompt Injection}
\label{sec:prmpt-injct-leak}
Prompt Injection in LLMs, including both injection and leaking, is a category of adversarial attacks that allow adversaries to hijack a model's output or even expose its training data. In the case of prompt injection, an adversary strategically crafts inputs, exploiting the model’s inherent biases or knowledge, to elicit specific or misleading outputs. Prompt leaking, a more specialized form of this attack, involves querying the model in such a way that it regurgitates its own prompt verbatim in its response. A common prompt injection strategy is to attempt to trick LLMs by adding triggers—common words, uncommon words, signs, sentences, etc.—into the prompt. To attack the few-shot examples, for instance, \textit{advICL} \cite{Wang2023AdversarialDA} leverages word-level perturbation, such as character insertion, deletion, swapping, and substitution to achieve this.  Kang et al. \citep{Kang2023ExploitingPB} dive deeper into the realm of LLMs, particularly focusing on models proficient in following instructions e.g., ChatGPT. They highlight an ironic twist: \textit{the enhanced instruction-following capabilities of such models inadvertently increase their vulnerability}. When strategically crafted prompts are presented to these LLMs, they can generate harmful outputs, including hate speech or conspiracy theories. This observation introduces a new layer of complexity to the security landscape surrounding LLMs, suggesting that their advanced capabilities may also be their main vulnerability as well. Greshake et al. \cite{Greshake2023NotWY} have brought to light a novel threat: ``Indirect Prompt Injection''.  In this scenario, adversaries cleverly embed prompts into external resources that LLMs access, such as websites. This method represents a departure from the traditional direct interaction with LLMs, as it allows attackers to exploit them remotely. Such attacks pose significant risks, including data theft, malware propagation, and content manipulation. This revelation underscores a critical shift in the approach to LLM exploitation, expanding the landscape of potential vulnerabilities. 

\paragraph{Prompt Leaking}
In a similar vein, Perez et al. \citep{Perez2022IgnorePP} focus on a specific aspect of prompt manipulation i.e., prompt leaking. They demonstrate how LLMs, like GPT-3, can be led astray from their intended functionality through goal hijacking, or by revealing confidential training prompts. Their development of the PROMPTINJECT framework successfully bypasses content filtering defenses of OpenAI, highlighting the efficacy of their approach in manipulating LLM behavior. Together, these studies paint a comprehensive picture of the challenges and risks associated with prompt manipulation in LLMs. They highlight the need for a nuanced understanding of both the technical capabilities and potential vulnerabilities of these advanced AI systems, emphasizing the importance of developing robust security and privacy solutions.

\paragraph{Mitigation Strategies} 

Techniques such as asking or eliminating each token in the prompt and assessing its effect on subsequent tasks, are among common detection strategies~\cite{ribeiro2016should,qi2020onion}. Another promising mitigation strategy is to decrease the negative impact of triggers is to filter outlier tokens that cause performance degradation~\cite{Xu2022ExploringTU}. Suo et al.~\cite{suo2024signedprompt} propose a method called \textit{Signed-Prompt}  which prevent prompt injection attacks in applications that utilize LLMs. Signed-Prompt utilizes digitally signed instructions by the authorized source to verify the source of instructions provided to the LLM, ensuring that the instructions come from a trusted and authenticated entity. Validating and sanitizing user input to prevent malicious instructions, considering the context in which instructions are given and filtering out instructions that deviate from the expected context, and dynamically adjusting defense mechanisms based on system behavior are among other mitigation strategies~\cite{liu2023prompt}.


\section{Altering the Behavior of LLMs via Model Editing}
\vspace{-4ex}
LLMs have become widely popular in various fields. However, a significant challenge arises when certain LLMs have an extensive number of parameters—often in the billions. The pressing question is: How can we mitigate undesirable behaviors, such as generating offensive content or providing incorrect answers, without requiring a complete retraining of the LLM? The answer to this inquiry lies in model editing, which refers to the process of modifying architecture, parameters, weights or other aspects of LLMs in order to improve their behavior, gradient editing, weight editing, memory-based editing and ensemble editing are common practices for model editing. Do note that this approach needs access to the parameters of the model so black-box models may not be able to be edited in this manner. In this section we dig deeper into editing methods.

\subsection{Gradient Editing} 
\vspace{-2ex}
Gradient Editing is a method that allows post-hoc modifications to pre-trained LLMs. As models grow in size, even initially accurate predictions can lead to errors as time progresses. Detecting all such failures during training is an impossible task. Therefore, enabling developers and end users to correct inaccurate outputs while preserving the overall model integrity is highly desirable. \par Model Editor Networks with Gradient Decomposition (MEND)~\cite{mitchell2021fast} is an example of gradient editing where an LLM is edited in order to achieve desirable behaviors. MEND involves training a set of Multi-Layer Perceptrons (MLPs). These MLPs modify gradients in a way that ensures local parameter edits do not negatively impact the model’s performance on unrelated inputs. The MEND process consists of two stages: training and subsequent editing. Researchers apply this method to T5, GPT, BERT, and BART models, evaluating its effectiveness on datasets such as zsRE Question-Answering, FEVER Fact-Checking, and Wikitext Generation. MEND effectively edits the largest available transformers, surpassing other methods in terms of the degree of modification.

\subsection{Weight Editing} 
\vspace{-2ex}
Weight Editing in the context of model editing involves modifying the weights (parameters) of a pre-trained LLM.~\cite{ilharco2023editing}. Rank-One Model Editing (ROME) proposed by Meng et al.~\cite{meng2022locating} is an example of weights editing that involves modifying the feedforward weights within a neural network to evaluate factual association recall. The approach scrutinizes neuron activations and adjusts weights to detect changes related to factual information. In this work a dataset of counterfactual assertions a.k.a COUNTERFACT is curated to assess counterfactual edits in LLMs. By employing causal tracing, they pinpoint the most critical MLP modules for retaining factual details. This work highlights the importance of middle layers in MLP modules for recalling factual information. Effective weight editing requires a crucial understanding of the model’s specific storage locations for information. This knowledge significantly impacts decisions during the editing process, guiding where modifications should occur. Specifically, empirical evidence indicates that factual information tends to be concentrated in the middle layers of the model~\citep{meng2022locating,meng2022mass}, whereas commonsense knowledge typically resides in the early layers~\cite{gupta2023editing}. 
\par Another example is a work by Ilharco et al.~\cite{ilharco2023editing}, where weight editing is practiced by creating task vectors. These vectors are constructed by subtracting the weights of a pre-trained model from the weights of the same model after fine-tuning it for a specific task. These task vectors represent directions within the weights space of the model. By combining and adjusting these task vectors using arithmetic operations (such as negation and addition), we can influence the behavior of the resulting model, ultimately enhancing its performance on specific tasks.

\subsection{Memory-based Model Editing}
\vspace{-2ex } Memory-Based Model Editing is a technique used to enhance LMs by making local updates to their behavior. In this context, model editors modify pre-trained base models to inject updated knowledge or correct undesirable behaviors. Their goal is to improve model performance by adjusting specific aspects of the model’s behavior, such as its predictions or responses.
\par One notable approach is SERAC~\cite{pmlr-v162-mitchell22a}, which leverages external memory to enhance model behavior. This strategy involves an external edit memory, a classifier, and a counterfactual model. Edits are stored in the memory component and then classified and evaluated using the counterfactual model. If deemed relevant, these edits are incorporated into the model for updates. The approach has been evaluated using T5-large, BERT, and BB-90M models on datasets such as question answering (QA), challenging QA (QA-hard), fact-checking (FC), and conversational sentiment (ConvSent), demonstrating remarkable success.
\par Another prominent method is Mass-Editing Memory in a Transformer (MEMIT)~\citep{meng2022mass}  which focuses on enhancing LLMs by incorporating additional memories (associations) that can scale to a large size. The primary objective of MEMIT is to modify the factual associations stored within the weights of LLMs. While ROME edits the LLM on a single basis, MEMIT has the capability to scale up to thousands of associations (memories) for models like GPT-J and GPT-NeoX. Furthermore, MEMIT enables updates to be made across multiple layers’ parameters. 
\par Gupta et al.~\cite{gupta2023editing} propose MEMIT$_{CSK}$ by 
extending the MEMIT framework to adapt it for handling commonsense knowledge. while MEMIT focuses on editing LMs to assess whether they store associations related to encyclopedic knowledge, MEMIT$_{CSK}$ specifically targets commonsense knowledge that differs from encyclopedic knowledge. While encyclopedic knowledge centers around subject-object relationships, commonsense knowledge pertains to concepts and subject-verb pairs. MEMIT$_{CSK}$, effectively corrects commonsense mistakes and can be applied to editing subjects, objects, and verbs. Through experiments, it is demonstrated that commonsense knowledge tends to be more prevalent in the early layers of the LM, in contrast to encyclopedic knowledge, which is typically found in the middle layers.

\subsection{Ensemble Editing and Beyond}
\vspace{-2ex}
Ensemble Editing refers to combining the power of multiple model editing techniques to create a more robust and effective approach. Wang et al.\citep{wang2023easyedit}, for instance, propose \textit{EasyEdit}, a framework for model editing, that
ensures user-friendly applicability across various LLMs. EasyEdit applies model editing techniques to assess specific hyperparameters such as certain layers or neurons. Customizable evaluation metrics are then employed to evaluate the effectiveness of the model editing. The framework’s success is demonstrated across several LLMs, including T5, GPT-J, GPT-NEO, GPT-2, LLaMA, and LLaMA-2, leveraging methods such as ROME, MEMIT, MEND, and others.

\par Despite all the aforementioned works on model editing, Yao et al.\citep{yao2023editing} conduct an analysis to investigate the performance of model editing methods. They introduce a novel dataset specifically designed for this purpose. Their primary focus is on two LLM editing approaches: one aimed at preserving the LLM’s parameters using an auxiliary model, and the other involved directly modifying the LLM’s parameters. Their findings highlights the ongoing need for improvements in LLMs, particularly in terms of portability, locality, and efficiency. As the field progresses, it becomes essential to enhance model editing techniques to boost performance of LLMs.

In addition, recent research explores the impact of model editing on language modeling. Hazra et al.~\cite{hazra2024sowing} find that maintaining factual correctness during model editing may compromise the model, leading to unsafe behaviors. Conversely, Wang et al.~\cite{wang2024detoxifying} propose an approach called DINM that detoxifies language models, ensuring safe content generation. As model editing is a relatively new area, further research is needed to assess its effectiveness in mitigating vulnerabilities mentioned earlier.

\section{Chroma Teaming: Uniting Color Teams Against Security Threats}
\vspace{-3ex}
\label{sec:red-green}
In this section, we introduce the concept of the ``Chroma Teaming'' which represents the collaborative synergy among red, blue, green, and purple teams in the field of LLM security. These teams work harmoniously toward a shared goal: fortifying LLMs against adversarial attacks.
\subsection{Red and Blue Teaming}
\vspace{-2ex}
 Red and blue teaming have been synonymous terms for testing security vulnerabilities. Red teams simulate cyberattacks to identify vulnerabilities, while blue teams focus on defense and prevention. With the advent of LLMs, a new area of research has emerged that leverages the red teaming approach from cybersecurity to assess adversarial capabilities that could affect LLMs. It is important to note that while some researchers consider jailbreaking and red teaming to be synonymous, we distinguish between these terms in our context. Jailbreaking has broader implications, often related to privacy concerns, whereas red teaming aims to demonstrate potential harms. 

\par The motivation behind red teaming lies in its ability to reveal the undesirable effects that LLMs can produce. These undesirable outputs span a wide range, including hate speech, violent statements, and even sexual content~\cite{Ganguli2022RedTL,Ge2023MARTIL}. To illustrate this, consider an example of red teaming: deliberately crafting a prompt to incite the LLM to generate violent content. If the prompt successfully bypasses the model’s defenses, the red teaming objective is achieved. Essentially, red teaming aims to assess which prompts or strategies can circumvent an LLM’s safeguards and lead to undesirable responses. 
\par Numerous studies have explored red teaming in the context of LLMs~\cite{Ge2023MARTIL,Perez2022RedTL,Bhardwaj2023RedTeamingLL}, shedding light on their strengths and weaknesses. Given their significant effectiveness, red teaming plays a pivotal role in understanding the potential adverse impacts of LLMs. For instance, Zhuo et al.~\cite{zhuo2023red}, have investigated whether ChatGPT produces hazardous outputs by employing prompt-injection techniques, while other studies~\cite{shi2023red,casper2023explore,perez2022red}, focus on specific red-teaming aspects including developing toxicity classifiers or using red-teaming LLMs to identify risky generations that might otherwise go unnoticed. A study conducted by Ganguli et al.~\cite{Ganguli2022RedTL} illustrates how different sampling techniques can deter specific elements of red teaming. For instance, Reinforcement Learning with Human Feedback (RLHF) has been shown to exhibit greater resilience compared to rejection sampling. However, these methods often require significant human involvement. To alleviate this manual burden, other researchers have explored ways to automate red teaming. For example, Lee et al.~\cite{lee2023query} utilize Bayesian optimization to conduct red teaming with fewer queries and reduced reliance on human assistance.
\subsection{Green and Purple Teaming}
\vspace{-2ex}
Interestingly, there is emerging research on a concept called “Green Teaming”~\cite{stapleton2023seeing}.Unlike red teaming, which primarily aims to uncover vulnerabilities and risks, green teaming investigates scenarios where seemingly unsafe content could still serve beneficial purposes. It acknowledges the nuanced situations where language models generate content that might be considered unsafe but has a specific intent. For instance, using language models to create intentionally flawed code for educational purposes falls within this category.
 \par As green teaming gains recognition, new variations of color teaming have emerged alongside red teaming. One such variation is purple teaming, as explored by Bhatt et al.~\cite{bhatt2023purple}. Their focus is on detecting insecure code generated by LLMs. Bhatt et al. provide a benchmark to assess the cybersecurity safety of LLMs, drawing from both red and blue teaming concepts. 
 \subsection{Rainbow Teaming and Beyond}
 \vspace{-2ex}
Going beyond the concept of red teaming, novel teaming approaches have emerged, including rainbow teaming~\cite{samvelyan2024rainbow}. Rainbow teaming builds upon the concept of red teaming, with a primary focus on generating a diverse set of adversarial prompts. While red teaming prioritizes success rates, rainbow teaming goes beyond and emphasizes on both success and diversity. It achieves this by utilizing feature descriptors and a mutation operator to create new adversarial prompts.

\par As time progresses, further advancements in color teaming will lead to improved strategies for mitigating adversarial attacks. Additionally, new perspectives or influences may emerge, as demonstrated by rainbow teaming. Observing these trends is crucial, especially when dealing with different or multiple modalities for LLMs. Summarily, it is crucial to utilize the unique strengths of different teaming strategies via ``Chroma Teaming'' where red teaming helps identify vulnerabilities, blue teaming safeguards against threats, green teaming generates innovative solutions, and purple teaming combines insights from both red and blue approaches to enhance overall resilience.

\section{Future Directions}
\vspace{-3ex}
This research paper offers a comprehensive study of the LLM vulnerabilities and effective strategies including chroma teaming and model editing to mitigate them.  However, as we explored throughout the paper, there exist limitations and challenges associated with the current strategies, which present opportunities for further advancements in the field of LLM security, vulnerability, and risk mitigation. These opportunities include but not limited to: (1) examining LLMs vulnerability at both model architecture and model size level, (2) investigating the role of transfer learning
and fine-tuning in model vulnerabilities, (3) identifying and mitigating emerging attacks, (4) evaluating the impact of attacks on specific models, (5) designing automated systems to reduce
human dependency in color teaming, (6) further exploration of model editing study across diverse datasets and model aspects e.g., (7) layers and parameters, (8) and developing unified platform to test multiple model editing methods.

\section{Conclusion}
\vspace{-3ex}
In this paper, we investigated three areas of LLM vulnerabilities, which impact various components within the LLM pipeline during different life cycle phases, namely training and inference. We also examine how each vulnerability affects LLMs, identifying effective strategies to mitigate them. Specifically, we delve deep into several color teaming and model editing strategies, as they can be applied to a broad spectrum of LLMs throughout their life cycle. Finally, we identify limitations of the current mitigation strategies and propose new research directions to bridge the existing gaps. We envision that the body of work contained in this paper will serve as a blueprint for further research in the security study of LLMs.

\newpage
\bibliography{colm2024_conference}
\bibliographystyle{colm2024_conference}

\appendix

\end{document}